\documentclass[letterpaper]{article} 
\usepackage{aaai2026}  
\usepackage{times}  
\usepackage{helvet}  
\usepackage{courier}  
\usepackage[hyphens]{url}  
\usepackage{graphicx} 
\usepackage{natbib}  
\usepackage{caption} 
\usepackage{algorithm}
\usepackage{algorithmic}

\usepackage{subfigure}
\usepackage{amsmath}
\usepackage{amssymb}
\usepackage{multirow}
\usepackage{booktabs}
\usepackage{utfsym}
\usepackage{colortbl}
\usepackage{newfloat}
\usepackage{listings}
\DeclareCaptionStyle{ruled}{labelfont=normalfont,labelsep=colon,strut=off} 
\floatstyle{ruled}
\newfloat{listing}{tb}{lst}{}
\floatname{listing}{Listing}
\title{Cross Modal Fine-Grained Alignment via\\ Granularity-Aware and Region-Uncertain Modeling}

\author{
    Jiale Liu\textsuperscript{\rm 1}\equalcontrib,
    Haoming Zhou\textsuperscript{\rm 1}\equalcontrib,
    Yishu Liu\textsuperscript{\rm 2},
    Bingzhi Chen\textsuperscript{\rm 3},
    Yuncheng Jiang\textsuperscript{\rm 1}\thanks{Corresponding author.}
}
\affiliations{
    \textsuperscript{\rm 1}South China Normal University, Guangzhou, China\\
    \textsuperscript{\rm 2}Harbin Institute of Technology, Shenzhen, China\\
    \textsuperscript{\rm 3}Beijing Institute of Technology, Zhuhai, China\\
    liujiale, jiangyuncheng@m.scnu.edu.cn, zhouhaoming.smile@gmail.com, 
    liuyishu@stu.hit.edu.cn, chenbingzhi@bit.edu.cn
}

\usepackage{bibentry}

\begin{document}

\maketitle

\begin{abstract}
Fine-grained image-text alignment is a pivotal challenge in multimodal learning, underpinning key applications such as visual question answering, image captioning, and vision-language navigation. Unlike global alignment, fine-grained alignment requires precise correspondence between localized visual regions and textual tokens, often hindered by noisy attention mechanisms and oversimplified modeling of cross-modal relationships. In this work, we identify two fundamental limitations of existing approaches: the lack of robust intra-modal mechanisms to assess the significance of visual and textual tokens, leading to poor generalization in complex scenes; and the absence of fine-grained uncertainty modeling, which fails to capture the one-to-many and many-to-one nature of region-word correspondences. To address these issues, we propose a unified approach that incorporates significance-aware and granularity-aware modeling and region-level uncertainty modeling. Our method leverages modality-specific biases to identify salient features without relying on brittle cross-modal attention, and represents region features as a mixture of Gaussian distributions to capture fine-grained uncertainty. Extensive experiments on Flickr30K and MS-COCO demonstrate that our approach achieves state-of-the-art performance across various backbone architectures, significantly enhancing the robustness and interpretability of fine-grained image-text alignment.
\end{abstract}

\begin{links}
    \link{Code}{https://github.com/H3IIoWorld/GRM}
\end{links}

\section{Introduction}
Fine-grained image-text alignment is a challenging and essential task in multimodal learning which has significant applications in tasks like visual question answering\cite{antol2015vqa,yu2025fine}, image captioning\cite{huang2019attention,zeng2025zero}, and vision-language navigation\cite{anderson2018vision,chen2025affordances}. Unlike global image-text alignment, which considers the overall semantic similarity between an image and a text, fine-grained image-text alignment demands detailed alignment between localized visual representations and the corresponding textual tokens. Achieving this fine-grained alignment is challenging due to the need to reason over compositional details such as object attributes, spatial relations, and localized entities, which requires more precise modeling of the visual-textual correspondences.\\
Despite the remarkable advances in cross-modal alignment, existing methods\cite{wu2025relation,zhao2024unifying,qiu2024mining} face two critical limitations that hinder their ability. First, while many approaches leverage cross-attention or fusion modules to model token-level interactions, the attention weights are typically driven by the retrieval objective. These attention maps are often noisy, lacking semantic grounding, and frequently focus on visually salient but semantically irrelevant regions. This results in a failure to explicitly identify which regions and tokens are truly critical for alignment, particularly in complex and ambiguous visual scenes. Without an effective mechanism to pinpoint the key areas for alignment, existing models struggle with misalignment, thereby impairing performance.\\
Second, uncertainty modeling has become increasingly recognized as crucial for capturing the inherent ambiguity in multimodal data. However, most recent methods treat uncertainty at the image-text pair level, assuming a one-to-one correspondence between image and text. In reality, the alignment is more nuanced: a single textual phrase may correspond to multiple regions, or a region may ambiguously match multiple text tokens. Yet, region-level uncertainty—capturing these one-to-many and many-to-one relationships—has been largely underexplored. This gap limits the ability of current methods to handle fine-grained uncertainty, which is critical for achieving robust and accurate cross-modal alignment.\\
Motivated by the limitations mentioned above, we raise the first fundamental question: \textit{How can we effectively model the significance of embedding tokens within each modality?} Some existing approaches directly apply cross-modal attention to identify salient tokens~\cite{laps}. However, such methods are often inefficient and overly dependent on specific image-text pairs, which compromises generalizability. 
Therefore, we argue that significance modeling should be performed intra-modally, leveraging the inherent statistical biases within each modality rather than depending on cross-modal interactions. This modality-specific modeling can better adapt to diverse and complex downstream tasks.\\
After obtaining modality-specific significance features, we further raise a second fundamental question: \textit{How can we model fine-grained uncertainty in the alignment process?} Inspired by the concept of probabilistic Distributional Representations, we adopt Gaussian distributions~\cite{chang2020data,ji2023map} to work out this problem. Specifically, we assume that each region feature can be represented as a multivariate Gaussian distribution, where the variance reflects the intrinsic uncertainty of the representation. To capture fine-grained uncertainty, we further model the image as a mixture of Gaussian distributions across multiple regions.\\
Motivated by these two insights, we propose a unified approach \textbf{GRM} (Cross Modal Fine-Grained Alignment via \textbf{G}ranularity-Aware and \textbf{R}egion-Uncertain \textbf{M}odeling) that integrates granularity-aware significance modeling and region-level uncertainty modeling to enhance cross-modal alignment. Specifically, the main contributions of this work are threefold:
\begin{itemize}
    \item We propose a Significance-aware and Granularity-aware Adapting approach that explicitly models intra-modal data biases to identify salient modality-specific features, thereby suppressing redundant information and enhancing alignment generalization.
    \item We introduce a prompt-driven mechanism for end-to-end region proposal, followed by the use of a mixture of Gaussians to model fine-grained region-level uncertainty, which improves the robustness of cross-modal alignment.
    \item We comprehensively evaluate our approach against existing fine-grained methods across diverse backbone architectures, consistently surpassing state-of-the-art performance on two benchmarks: Flickr30K and MS-COCO.
\end{itemize}

\section{Related Work}
\subsection{Fine-grained Cross-modal Alignment}
Fine-grained cross-modal alignment aims to establish precise correspondences between localized visual regions and textual elements, serving as a foundational task in multimodal understanding. Recent research~\cite{faghri2017vse++,scan,sgr,chan,laps,avse,cora,hrem} in this area typically follows a dual-stream architecture that separately encodes images and texts before aligning them through cross-modal interactions.
Existing methods can be broadly categorized based on their visual encoders. One line of work relies on two-stage object detectors~\cite{cora,hrem,qin2022deep} (e.g., Faster R-CNN~\cite{fasterrcnn}) to extract region-level features as inputs to the alignment module. While such approaches offer explicit localization cues, they suffer from several drawbacks: the reliance on pre-trained detection models introduces error propagation and limits flexibility; in addition, adapting to downstream tasks often requires additional annotations and retraining, which increases the system's complexity and domain sensitivity.
In contrast, a more recent trend adopts end-to-end transformer-based vision encoders, particularly Vision Transformers (ViT)~\cite{dosovitskiy2020image,liu2021swin}, to directly model visual features without explicit region proposals. This design benefits from unified optimization, reduced dependence on external annotations, and better scalability across datasets. Motivated by these advantages, we adopt a ViT-based architecture in our framework.
Although CLIP-style~\cite{lu2022cots,wang2023agree,jiang2023cross,zhang2024user,zheng2023make} pretraining has proven effective for global image-text matching, its representation is typically optimized for coarse-level similarity and lacks the granularity required for fine-grained alignment. Therefore, we opt for independently pre-trained ViT and BERT~\cite{devlin2019bert} models to retain modality-specific inductive biases and allow greater flexibility in downstream fine-grained alignment tasks.

\subsection{Uncertainty Quantification}
Uncertainty modeling has recently gained attention in multimodal learning due to its potential to handle ambiguity, noise, and diverse interpretations inherent in image-text data. Prior works have explored uncertainty estimation through various paradigms, such as Bayesian neural networks~\cite{kingma2015variational,molchanov2017variational,kohl2018probabilistic}, probabilistic embeddings~\cite{chen2022composed}, and distributional representations~\cite{gao2024embracing}.
Broadly speaking, existing approaches can be categorized into two types: (1) Pair-level uncertainty modeling~\cite{li2023prototype}, which estimates uncertainty at the global image-text pair level, often using confidence scores or variance-aware similarity metrics to account for ambiguity in the retrieval task. (2) Feature-level probabilistic modeling~\cite{chen2022composed,gao2024embracing}, which represents embeddings as probability distributions—typically Gaussians—to capture semantic variance and improve robustness in the embedding space. While these methods provide a useful first step toward modeling uncertainty, they predominantly focus on global or holistic representations, assuming a one-to-one alignment between an entire image and a sentence. This assumption fails to capture the fine-grained one-to-many and many-to-one relationships between visual regions and textual tokens, which are critical in tasks like phrase grounding and referring expression comprehension. As a result, global uncertainty modeling often overlooks localized ambiguities and leads to suboptimal alignment in complex scenes.

\begin{figure*}[t]
\centering
\includegraphics[width=1.0\linewidth]{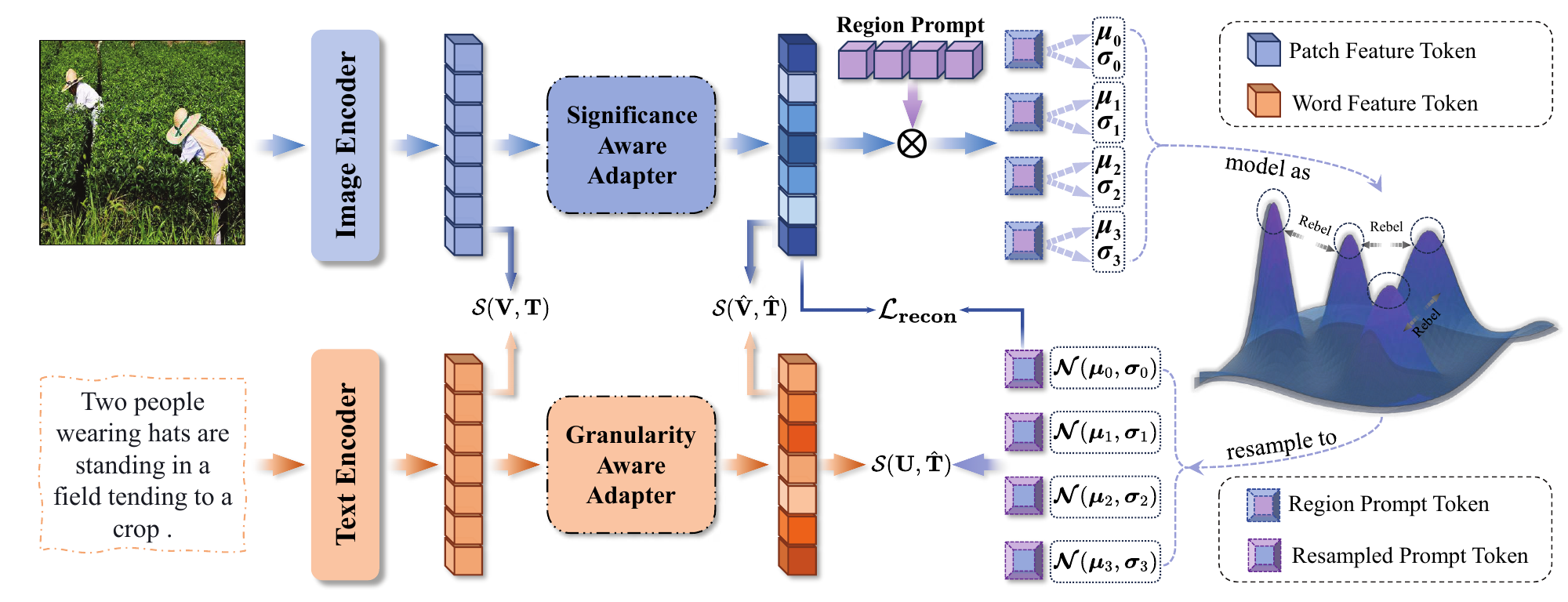}
\caption{
The Overview of the Proposed GRM.The visual encoder $f_v(\cdot)$ and text encoder $f_t(\cdot)$ independently encode input image and text instances to obtain their respective representations, $\mathbf{V}$ and $\mathbf{T}$. These embeddings are then passed through two structurally identical but functionally distinct adapters: the \textbf{Significance-aware Adapter} and the \textbf{Granularity-aware Adapter}, which learn modality-specific distribution biases. Subsequently, the image embeddings undergo \textbf{region-level prompt learning} and \textbf{uncertainty modeling} to capture fine-grained semantic variations. Finally, a \textbf{multi-level alignment strategy} is applied to effectively align the cross-modal knowledge between images and texts.
}

\label{fig:model}
\end{figure*}
\section{Method}
\label{sec:method}
The overview of our GRM is illustrated in Figure~\ref{fig:model}. We first introduce the architecture of the modality encoders, followed by the detailed implementation of the Significancy-aware and Granularity-aware Adapters. Finally, we present the methodological details of region prompting and fine-grained uncertainty modeling.

\subsection{Dual-encoder Feature Extraction}
\label{sub:feature_extractor}
Our method adopts a dual-encoder architecture, wherein visual and textual features are independently extracted by modality-specific encoders without shared parameters.
Given an input image instance $I$, we employ vision transformers~\cite{dosovitskiy2020image,liu2021swin} as visual encoder $f_v(\cdot)$ to obtain the initial visual representation $\mathbf{V}=f_v(I)\in\mathbb{R}^{L_v\times d}$, where $v_i$ denotes the $i$-th patch token generated by patch embedding of the visual encoder. Similarly, for a textual instance $T$, we utilize BERT~\cite{devlin2019bert} as textual encoder $f_t(\cdot)$ to obtain the corresponding textual representation $\mathbf{T}=f_t(T)\in\mathbb{R}^{L_t\times d}$, where $t_i$ denotes the $i$-th word token generated by the tokenizer of the textual encoder. $L_v$ and $L_t$ denotes the length of image tokens and text tokens, and $d$ indicates the latent dimension.

\begin{figure}[t]
\centering
\includegraphics[width=1.0\linewidth]{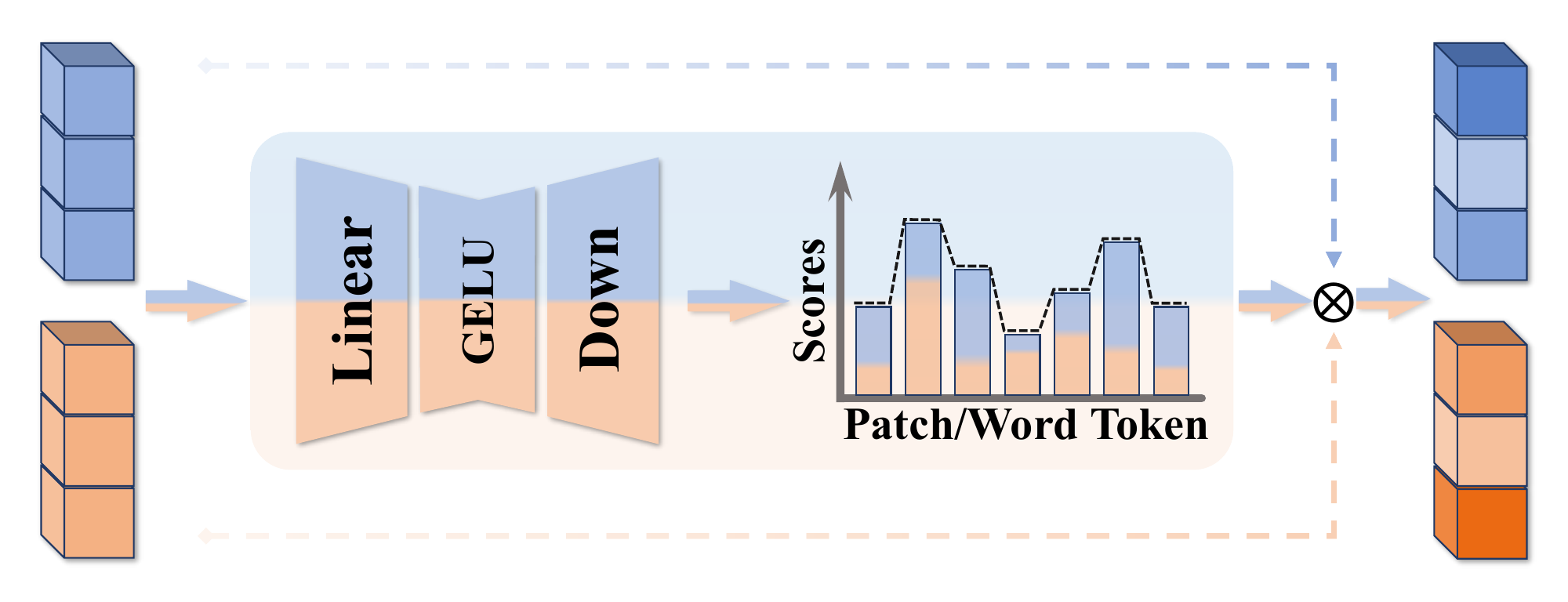}
\caption{
The detailed architecture of the Significance-aware Adapter and the Granularity-aware Adapter.}

\label{fig:adapter}
\end{figure}

\subsection{Significancy-aware and Granularity-aware Adapting}
\label{sub:adapters}
The architectures of the Significance-aware Adapter and Granularity-aware Adapter are illustrated in Figure~\ref{fig:adapter}. While both modules adopt an identical architectural design, they are instantiated independently and operate on different modalities, thereby fulfilling distinct functional roles. This modality-specific divergence, despite structural symmetry, reveals an intriguing dynamic that we empirically validate in the experimental section. Given the shared architecture between the two adapters, we present the detailed formulation using the visual input as a representative example in the following discussion. Given the visual representation $\mathbf{V} \in \mathbb{R}$ the feature adapting process is formulated as follows:
\begin{equation}
\begin{split}
\mathbf{D}_v =& \mathbf{W}_2^{\mathbf{V}}(\psi (\mathbf{W}_1^\mathbf{V}\mathbf{V})) \in \mathbb{R}^{L_v\times 2}
\\
\mathbf{A}_\mathbf{V}& = f_g(\mathbf{D}_v, \tau)[:,1] \in [0,1]^{L_v}
\\
&\hat{\mathbf{V}} = \mathbf{M} \odot \mathbf{A}_\mathbf{V} \otimes \mathbf{1}_d
\end{split}
\end{equation}
Here, $\odot$ and $\otimes$ denotes element-wise multiplication and the outer product, $\psi$ and $f_g$ indicates the activation function (GELU in our implementation) and the Gumbel-Softmax function respectively. And $\mathbf{1}_d$ is a $d$-dimensional all-one vector. The temperature parameter $\tau$ is a scalar that controls the sharpness of the selection distribution.

\subsection{Region Prompting and Uncertainty Modeling}
\label{sub:uncertainty}
While prior studies~\cite{gao2024embracing,li2023prototype} have primarily modeled uncertainty at the holistic image-text level and achieved competitive performance, such formulations overlook the intrinsic granularity of cross-modal alignment. In contrast, we advocate for a fine-grained perspective, positing that matching uncertainty should be decomposed across localized image regions and individual textual tokens. This paradigm shift enables a more nuanced characterization of semantic ambiguity and forms the basis for our proposed uncertainty-aware fine-grained matching approach.
\subsubsection{Region Prompting.} 
Inspired by prompt learning\cite{vpt,pei2024sa2vp,wang2024cross}, we first introduce a set of learnable prompts $\mathbf{P}=\{p_0,...,p_{K-1}\}\in\mathbb{R}^{K\times d}$ to model the latent region representations within the image, where each prompt $p_i$ serves as $i$-th semantic proxy for a potential image region. We then transfer visual knowledge from image representations to these region prompts, thereby generating region-aware prompt embeddings with regional semantic information. To ensure scale invariance across queries, we apply L2 normalization for $\mathbf{P}$ to get $\hat{\mathbf{P}}=\{\hat{p}_0,...,\hat{p}_{K-1}\}\in\mathbb{R}^{K\times d}$. Subsequently, we compute pairwise attention scores between each image patch token $v_i$ and every region prompt $p_k$ to capture their semantic affinity. Formally,
\begin{equation}
\label{patch-prompt attention}
\begin{split}
    &\mathbf{A}_r = \psi (\hat{\mathbf{V}}\cdot \hat{\mathbf{P}}^\top)\in\mathbb{R}^{L_v\times K},\\
    \hat{\mathbf{A}}_r^{lk} &= \frac{\mathbf{A}^{lk}_r}{\sum_{l'=1}^{L_v} \mathbf{A}^{l'k}_r},\quad\hat{\mathbf{A}}_r\in\mathbb{R}^{L_v\times K}.
\end{split}
\end{equation}
Here, $\psi $ denotes the sigmoid function, as a single patch may be associated with multiple regions simultaneously. Finally, we leverage the normalized attention matrix $\hat{\mathbf{A}}_r^{lk}$ to softly aggregate patch features into the region prompts, and compute the $k$-th mean prompt representation enriched with region-level semantics. The equation is expressed as:
\begin{equation}
    \boldsymbol{\mu}_k = \sum_{l=1}^{L_v} \hat{\mathbf{A}}^{lk}_r \mathbf{\hat{\mathbf{V}}}^l, \boldsymbol{\mu} \in \mathbb{R}^{K \times d}.
\end{equation}

\begin{figure}[t]
\centering
\includegraphics[width=1.0\linewidth]{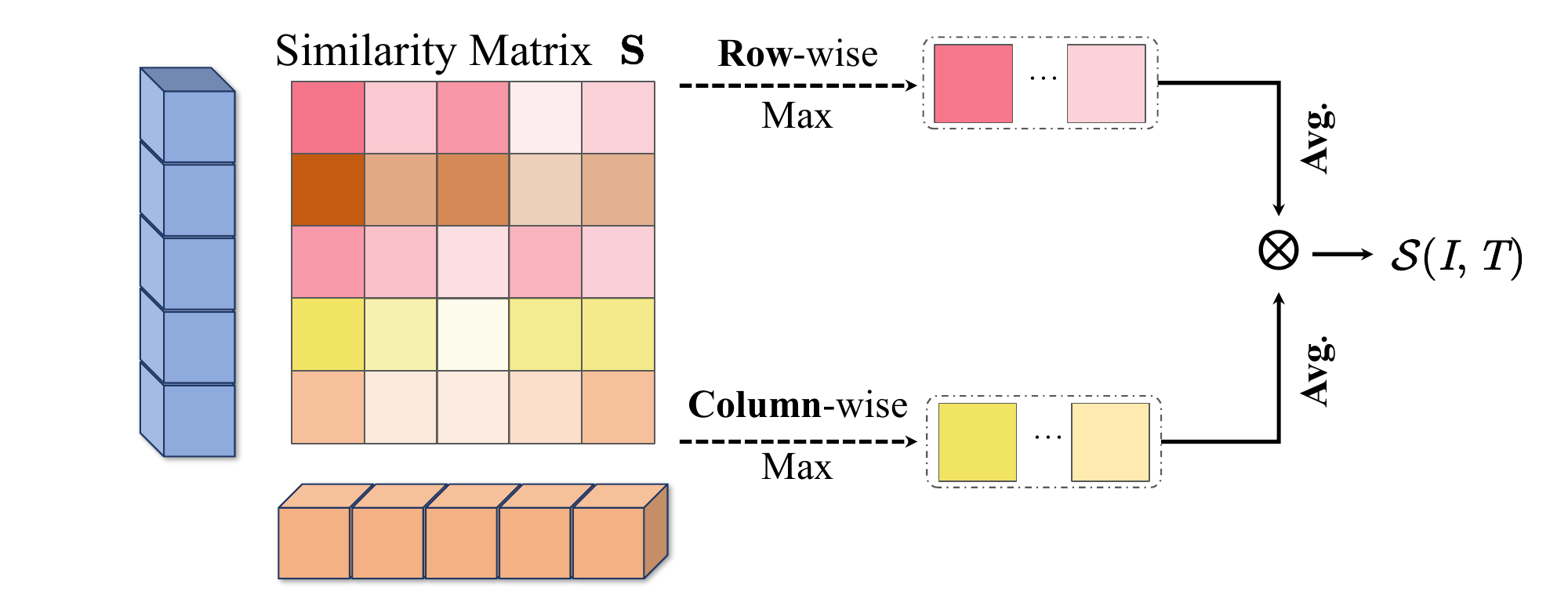}
\caption{
The computation process of bidirectional image-text alignment.
}

\label{fig:alignment}
\end{figure}
\subsubsection{Uncertainty Modeling:}
To enhance the expressiveness of region-level semantic modeling, we adopt a variational perspective~\cite{kingma2013vae} by representing the semantics of each region as a Gaussian distribution. This formulation explicitly captures the inherent ambiguity and uncertainty in the semantic aggregation process. Through a reparameterization-based sampling mechanism, token-level semantics are injected into region-level representations, thereby improving the model’s adaptability and generalization capability to semantic structures. \\
Specifically, we employ a learnable prediction network $\boldsymbol{\phi}$ to estimate the log-variance $\log \boldsymbol{\sigma}^2$ for each region prompt, which serves as an indicator of the model’s confidence in the corresponding semantic representation. Formally,
\begin{equation}
    \log \boldsymbol{\sigma}_k^2 = \boldsymbol{\phi}(\boldsymbol{\mu}_k), \quad \log \boldsymbol{\sigma}^2 \in \mathbb{R}^{K \times d}.
\end{equation}
We adopt the reparameterization trick to sample from each region’s distribution, thereby introducing diverse uncertainty and enhancing the expressiveness of region-level representations. Unlike conventional region-token embeddings that are treated as fixed vectors, we independently inject region-level noise into each patch–region pair to construct patch-aware sampled representations. Formally,
\begin{equation}
    \mathbf{z}_{lk} = \boldsymbol{\mu}_k + \boldsymbol{\epsilon}_{lk} \odot \exp\left(\frac{1}{2} \log \boldsymbol{\sigma}_k^2 \right), \quad \boldsymbol{\epsilon} \sim \mathcal{N}(0, \mathbf{I}).
\end{equation}
Here, $\epsilon$ denotes the noise sampled from a standard normal distribution $\mathcal{N}(0, \mathbf{I})$, where $\mathbf{I}$ refers to the identity covariance matrix. Then we apply the normalized attention matrix to perform a weighted aggregation of the sampled features $\mathbf{z}_{lk}$, resulting in an uncertainty-aware semantic representation for each region $\mathbf{u}_k$:
\begin{equation}
    \mathbf{u}_k = \sum_{l=1}^{L_v} \hat{\mathbf{A}}_{lk} \cdot \mathbf{z}_{lk}, \quad \mathbf{u} \in \mathbb{R}^{d}.
\end{equation}
Finally, we obtain the uncertainty-aware region representation $\mathbf{U}=\{ \mathbf{u}_0,\dots,\mathbf{u}_{K-1} \}\in \mathbb{R}^{K\times d}$.

\begin{table*}[ht!]
\renewcommand\arraystretch{.85}
\setlength{\tabcolsep}{1pt}
\footnotesize

\begin{tabular}{l|c|cccccc|c|cccccc|c|cccccc|cc}
\toprule
\multirow{3}{*}{Method} & \multirow{3}{*}{\textit{FG}} & \multicolumn{7}{c|}{Flick30k-1k}                                                                                       & \multicolumn{7}{c|}{MS-COCO 1k}                                                                                        & \multicolumn{7}{c}{MS-COCO 5k}                                                                                        &                      \\
                        &                     & \multicolumn{3}{c}{Image to Text}             & \multicolumn{3}{c|}{Text to Image}             & \multirow{2}{*}{rSum} & \multicolumn{3}{c}{Image to Text}             & \multicolumn{3}{c|}{Text to Image}             & \multirow{2}{*}{rSum} & \multicolumn{3}{c}{Image to Text}             & \multicolumn{3}{c|}{Text to Image}             & \multirow{2}{*}{rSum} & \multirow{2}{*}{}    \\
                        &                     & R@1           & R@5           & R@10          & R@1           & R@5           & R@10          &                       & R@1           & R@5           & R@10          & R@1           & R@5           & R@10          &                       & R@1           & R@5           & R@10          & R@1           & R@5           & R@10          &                       &                      \\
                        \midrule
\multicolumn{23}{l}{\textbf{\textit{Faster R-CNN + BERT-base}, 36 pre-computed regions}}                                                                                                                                                                                                                                                                                                                                                            & \multicolumn{1}{l}{} \\
\midrule
VSE++                   & \usym{1F5F4}          & 81.7 & 95.4          & 97.6          & 61.4          & 85.9          & 91.5          & 513.5                 & 79.7          & 96.4          & 98.9          & 64.8          & 91.4          & 96.3          & 527.5                 & 58.3          & 85.3          & 92.3          & 42.4          & 72.7          & 83.2          & 434.2                 &                      \\
MV-VSE                  & \usym{1F5F4}          & 82.1 & 95.8          & 97.9          & 63.1          & 86.7          & 92.3          & 517.5                 & 80.4          & 96.6          & 99.0            & 64.9          & 91.2          & 96.0             & 528.1                 & 59.1          & 86.3          & 92.5          & 42.5          & 72.8          & 83.1          & 436.3                               \\
CHAN                    & \usym{1F5F8}         & 80.6 & 96.1          & 97.8          & 63.9          & 87.5          & 92.6          & 518.5                 & 81.4          & 96.9          & 98.9          & 66.5          & 92.1          & 96.7          & 532.5                 & 59.8          & 87.2          & 93.3          & 44.9          & 74.5          & 84.2          & 443.9                                  \\
HERM                    & \usym{1F5F8}          & 84.0    & 96.1          & 98.6          & 64.4          & 88.0             & 93.1          & 524.2                 & 82.9          & 96.9          & 99.0             & 67.1          & 92.0             & 96.6          & 534.5                 & 64.0             & 88.5          & 93.7          & 45.4          & 75.1          & 84.3          & 451.0                                   \\
CORA                    & \usym{1F5F8}           & 83.4 & 95.9          & 98.6          & 64.1          & 88.1          & 93.1          & 523.3                 & 82.8          & 97.3          & 99.0             & 67.3          & 92.4          & 96.9          & 535.6                 & 64.3          & 87.5          & 93.6          & 45.4          & 74.7          & 84.6          & 450.1                                   \\
\midrule
\multicolumn{23}{l}{\textbf{\textit{ViT-Base-224 + BERT-base}, 14$\times$14 patches}}  &                   \multicolumn{1}{c}{} \\
\midrule
SCAN                    & \usym{1F5F8}           & 69.5 & 90.9          & 95.6          & 56.4          & 83.1          & 90.0             & 485.6                 & 76.0            & 95.4          & 98.1          & 64.5          & 90.8          & 95.8          & 520.6                 & 53.9          & 81.8          & 90.0             & 42.9          & 72.3          & 82.5          & 423.5                                  \\
VSE++                   & \usym{1F5F4}          & 71.8 & 92.8          & 96.5          & 59.4          & 84.7          & 90.9          & 496.1                 & 75.0             & 94.6          & 98.0             & 62.7          & 89.4          & 94.9          & 514.6                 & 52.4          & 80.3          & 88.8          & 40.6          & 70.4          & 81.1          & 413.4                                   \\
SGR                     & \usym{1F5F8}          & 69.7 & 90.8          & 95.2          & 59.1          & 84.1          & 89.9          & 488.7                 & 77.2          & 95.0             & 98.0             & 65.1          & 90.7          & 95.8          & 521.8                 & 54.9          & 82.8          & 90.5          & 42.8          & 72.2          & 82.5          & 425.8                                   \\
CHAN                    & \usym{1F5F8}           & 69.2 & 91.8          & 95.0             & 58.4          & 84.9          & 90.6          & 489.9                 & 77.1          & 95.1          & 98.1          & 65.0             & 91.0             & 96.0             & 522.2                 & 56.3          & 83.2          & 90.1          & 43.0             & 72.6          & 82.8          & 428.0                                      \\
LAPS                    & \usym{1F5F8}          & 74.0    & 93.4          & 97.4          & 62.5          & 87.3          & 92.7          & 507.3                 & 78.7          & 95.5          & 98.3          & 66.2          & 91.3          & 96.2          & 526.3                 & 57.5          & 84.0             & 90.8          & 44.5          & 74.0             & 83.6          & 434.4                                   \\
AVSE                    & \usym{1F5F4}         & 76.0 &	\textbf{94.6}	&97.5	&62.7	&88.4	&93.1	&512.3	&79.8	&95.6	&98.3	&67.0	&91.5	&96.3	&528.5	&58.8	&84.3	&91.0	&\textbf{45.1}	&74.3	&83.9	&437.4
              \\
\rowcolor{gray!20}\rowcolor{gray!20}Ours                    & \usym{1F5F8}  & \textbf{77.2}&	94.5	&\textbf{97.9}&	\textbf{64.6}&	\textbf{88.5}&	\textbf{93.5}	&\textbf{516.2}
        & \textbf{80.8} & \textbf{96.7} & \textbf{98.9} & \textbf{67.6} & \textbf{91.8} & \textbf{96.7} & \textbf{532.5}        & \textbf{60.1} & \textbf{86.7} & \textbf{93.2} & 44.4 & \textbf{74.5} & \textbf{84.1} & \textbf{443.0}            \\
\midrule
\multicolumn{23}{l}{\textbf{\textit{ViT-Base-384 + BERT-base}, 24$\times$24 patches}}     &                                       \multicolumn{1}{l}{} \\
\midrule
SCAN                    & \usym{1F5F8}           & 75.4 & 94.4          & 96.9          & 63.6          & 88.6          & 93.5          & 512.5                 & 77.0             & 95.7          & 98.4          & 64.6          & 91.1          & 96.2          & 523.0                    & 54.9          & 82.8          & 90.4          & 42.4          & 72.4          & 82.8          & 425.8                                   \\
VSE++                   & \usym{1F5F4}          & 77.1 & 95.7          & 97.5          & 65.8          & 90.2          & 94.3          & 520.5                 & 76.1          & 95.5          & 98.5          & 65.1          & 91.6          & 96.3          & 523.1                 & 53.3          & 81.8          & 90.0             & 42.6          & 72.6          & 82.9          & 423.1                                   \\
SGR                     & \usym{1F5F8}          & 76.9 & 94.9          & 98.1          & 64.2          & 88.4          & 93.3          & 515.8                 & 75.8          & 95.7          & 98.6          & 65.6          & 92.0             & 96.5          & 524.2                 & 53.3          & 81.0             & 89.6          & 42.9          & 73.1          & 83.7          & 423.6                                     \\
CHAN                    & \usym{1F5F8}         & 75.4 & 94.5          & 97.6          & 63.2          & 88.6          & 93.1          & 512.4                 & 78.1          & 95.8          & 98.6          & 66.1          & 92.1          & 96.6          & 527.3                 & 55.6          & 83.8          & 91.2          & 43.4          & 73.6          & 83.5          & 431.1                                   \\
LAPS                    & \usym{1F5F8}           & 79.0    & 96.0             & 98.1          & 67.3          & 90.5          & 94.5          & 525.4                 & 78.6          & 96.3          & 98.9          & 68.0             & 92.4          & 96.8          & 531.0                    & 57.4          & 84.9          & 92.5          & 46.4          & 75.8         & 85.2          & 442.2                                 \\
AVSE                    & \usym{1F5F4}      &    80.3 &	96.4 &	98.7 &	67.9 &	91.2 &	94.7 &	529.2 &	81.1 &	97.1 &	99.0 &	68.3 &	92.7 &	97.0 &	535.2 &	61.2 &	86.8 &	93.2 &	46.2 &	75.9 &	85.0 &	448.3
                                 \\
\rowcolor{gray!20}\rowcolor{gray!20}Ours                    & \usym{1F5F8}           &
\textbf{80.5}    & \textbf{97.5}    & \textbf{98.8}    & \textbf{68.5}    & \textbf{91.6}    & \textbf{94.9}    & \textbf{531.8}
            & \textbf{81.8}    & \textbf{97.7}    & \textbf{99.1}    & \textbf{69.4}    & \textbf{92.9}    & \textbf{97.3}    & \textbf{538.2}            & \textbf{62.1}    & \textbf{87.4}    & \textbf{93.4}    & \textbf{47.9}    & \textbf{76.5}    & \textbf{85.9}    & \textbf{451.2}                     \\
\midrule
\multicolumn{23}{l}{\textbf{\textit{Swin-base-224 + BERT-base}, 7$\times$7 patches}}                                 & \multicolumn{1}{l}{} \\
\midrule
SCAN                    & \usym{1F5F8}         & 79.0    & 95.9          & 98.2          & 67.7          & 90.6          & 94.9          & 526.3                 & 83.3          & 97.5          & 99.3          & 71.0             & 93.0             & 96.7          & 540.9                 & 64.0            & 88.2          & 94.2          & 49.9          & 78.0             & 86.6          & 460.9                                   \\
VSE++                   & \usym{1F5F4}          & 82.5 & 96.5          & 98.9          & 70.0             & 91.4          & 95.1          & 534.4                 & 80.9          & 97.0             & 99.1          & 69.7          & 93.1          & 97.1          & 536.9                 & 60.7          & 86.6          & 93.2          & 48.1          & 77.1          & 86.1          & 451.8                                  \\
SGR                     & \usym{1F5F8}          & 80.4 & 97.0             & 98.7          & 66.9          & 90.2          & 94.5          & 527.6                 & 81.2          & 97.1          & 99.1          & 69.9          & 93.2          & 97.2          & 537.7                 & 61.0             & 86.7          & 93.2          & 48.6          & 77.2          & 86.3          & 453.1                              \\
CHAN                    & \usym{1F5F8}           & 81.4 & 97.0             & 98.6          & 68.5          & 90.6          & 94.5          & 530.6                 & 81.6          & 97.2          & 99.3          & 70.6          & 93.7          & 97.6          & 539.8                 & 64.1          & 87.9          & 93.5          & 49.1          & 77.3          & 86.1          & 458.0                                 \\
LAPS                    & \usym{1F5F8}          & 82.4 & 97.4          & 99.5          & 70.0             & 91.7          & 95.4          & 536.3                 & 84.0            & 97.6          & 99.3          & 72.1          & 93.7          & 97.3          & 544.1                 & 64.5          & 89.2          & 94.4          & 51.6          & 78.9          & 87.2          & 465.8                                \\
AVSE                    & \usym{1F5F4}          & 83.9 & 97.4          & 99.4          & 70.0            & 92.4          & 95.6          & 538.7                 & 84.9          & \textbf{98.0 }   & 99.3 & 72.1          & \textbf{94.0 }   & 97.4 & 545.7                 & 66.2          & 89.8          & 94.7          & 51.7          & 79.2          & 87.3          & 468.9                            \\
\rowcolor{gray!20}\rowcolor{gray!20}Ours                    & \usym{1F5F8}          & \textbf{85.6}	& \textbf{98.7} &	\textbf{99.5} &	\textbf{73.0} &	\textbf{92.9} &	\textbf{96.2} &	\textbf{546.0}
        & \textbf{85.2} & 97.8 & \textbf{99.6} & \textbf{73.2} & 93.9 & \textbf{97.8} & \textbf{547.5}        & \textbf{66.3} & \textbf{90.1} & \textbf{94.9} & \textbf{52.1} & \textbf{79.8} & \textbf{87.6} & \textbf{470.8}                 \\
\midrule
\multicolumn{23}{l}{\textbf{\textit{Swin-base-384 + BERT-base}, 12$\times$12 patches}}                                         & \multicolumn{1}{l}{} \\
\midrule
SCAN                    & \usym{1F5F8}          & 81.9 & 96.9          & 98.9          & 70.0             & 92.7          & 95.8          & 536.1                 & 82.9          & 97.7          & 99.4          & 71.3          & 93.5          & 97.3          & 542.1                 & 63.0             & 88.5          & 94.3          & 50.1          & 78.9          & 87.4          & 462.2                                    \\
VSE++                   & \usym{1F5F4}         & 83.3 & 97.5          & 99.2          & 71.1          & 93.2          & 96.2          & 540.6                 & 81.6          & 96.8          & 99.1          & 69.1          & 92.7          & 96.7          & 536.1                 & 61.1          & 87.3          & 93.3          & 47.8          & 76.9          & 85.9          & 452.4                                     \\
SGR                     & \usym{1F5F8}          & 80.7 & 96.8          & 99.0             & 69.9          & 91.7          & 95.3          & 533.4                 & 81.9          & 96.7          & 99.1          & 69.3          & 92.8          & 96.7          & 536.6                 & 62.8          & 87.0             & 92.9          & 48.1          & 77.0             & 86.0             & 453.8                                  \\
CHAN                    & \usym{1F5F8}          & 81.2 & 96.7          & 98.8          & 70.3          & 92.2          & 95.9          & 535.0                   & 83.1          & 97.3          & 99.2          & 70.4          & 93.1          & 97.1          & 540.2                 & 63.4          & 88.4          & 94.1          & 49.2          & 77.9          & 86.6          & 459.5                                   \\
LAPS                    & \usym{1F5F8}           & 85.1 & 97.7          & 99.2          & 74.0            & 93.0             & 96.3          & 545.3                 & 84.1          & 97.4          & 99.2          & 72.1          & 93.9          & 97.4          & 544.1                 & 67.1          & 88.6          & 94.3          & 53.0            & 79.5          & 87.6          & 470.1                                    \\
AVSE                    & \usym{1F5F4}          & 87.1 & 98.3          & 99.2          & 73.6          & 93.5          & 96.5          & 548.2                 & \textbf{85.1}          & 98.2          & \textbf{99.5}          & 71.6          & 94.0             & 97.5          & 545.9                 & 68.6          & \textbf{90.2}          & 95.6          & 52.2          & 79.6          & 87.8          & 474.0                                     \\
\rowcolor{gray!20}\rowcolor{gray!20}Ours                    & \usym{1F5F8}
           & \textbf{87.7}    & \textbf{98.8}    & \textbf{99.5}    & \textbf{75.3}    & \textbf{93.5}    & \textbf{96.9}    & \textbf{550.7}            & 85.0    & \textbf{97.9}    & 99.4    & \textbf{73.5}    & \textbf{94.4}    & \textbf{97.8}    & \textbf{548.0}            & \textbf{69.4}    & 90.0    & \textbf{95.7}    & \textbf{54.4}    & \textbf{80.6}    & \textbf{88.2}    & \textbf{478.3}            \\
\bottomrule
\end{tabular}
\caption{Comparison of image-text retrieval performance on the Flickr30K and MS-COCO test sets. We list detailed information about feature encoders, image resolution and the number of regions/patches obtained by the visual encoder (e.g., ``ViT-Base-224'' denotes the base version of the Vision Transformer with a 224$\times$224 image resolution input, using 16$\times$16 pixel patches, resulting in 14$\times$14 visual patches per image). \textbf{\textit{FG}} indicates whether fine-grained cross-modal alignment is employed. The best results are highlighted in \textbf{bold}.}
\label{tab:main}
\end{table*}
\subsection{Multi-objective Optimization}
\label{sub:optimization}
\subsubsection{Multi-level and Bidirectional Image-Text Alignment:}
To comprehensively capture the semantic correspondences between images and texts, we introduce a multi-level, bidirectional token-wise alignment mechanism, which enables both fine-grained and semantically abstract interactions between the two modalities. We first define the token-level similarity between visual and textual features as $\mathbf{S}=\mathbf{T}\cdot\mathbf{V}^\top\in\mathbb{R}^{L_v\times L_t}$. To obtain a scalar instance-level similarity score $\mathcal{S}$ between the image and the caption, we perform bidirectional aggregation over $\mathbf{S}$. The computation process is illustrated in Figure~\ref{fig:alignment} and formally,
\begin{equation}
\label{sim}
    \mathcal{S}= \underbrace{\frac{1}{L_t} \sum_{l=1}^{L_t} \max_{L_v} \mathbf{S}_{l,L_v}}_{\text{Text2Image}}+\underbrace{\frac{1}{L_v} \sum_{l=1}^{L_v} \max_{L_t} \mathbf{S}_{l,L_t}}_{\text{Image2Text}}\in\mathbb{R}^{B\times B},
\end{equation}
where $B$ denotes the batch size.
For the original visual and textual feature pair $({\mathbf{T}, \mathbf{V}})$, the significance-aware and granularity-aware feature pair $({\hat{\mathbf{T}}, \hat{\mathbf{V}}})$, and the uncertainty-aware region-text pair $({\hat{\mathbf{T}}, \mathbf{U}})$, we obtain three levels of similarity maps—$\mathcal{S}^{\text{ori}}$, $\mathcal{S}^{\text{key}}$, and $\mathcal{S}^{\text{unc}}$—computed via the equation~\ref{sim}. We then compute the contrastive loss with hard negative mining~\cite{faghri2017vse++} for each of the three similarity matrices via equation~\ref{ContrastiveLoss}:
\begin{equation}
\label{ContrastiveLoss}
    \mathcal{L} = \sum_{i=1}^{B} \sum_{\substack{j=1 \\ j \ne i}}^{B} \left[ \alpha + \mathcal{S}_{ij} - S_{ii} \right]_+ + \left[ \alpha + \mathcal{S}_{ji} - \mathcal{S}_{ii} \right]_+,
\end{equation}
resulting in three levels of contrastive objectives $\mathcal{L}_{\text{con}}^{\text{ori}}$, $\mathcal{L}_{\text{con}}^{\text{key}}$, and $\mathcal{L}_{\text{con}}^{\text{unc}}$. Here, $\alpha$ denotes the margin hyperparameter, $S_{ij}$ represents the similarity between the $i$-th image and the $j$-th text, and $[x]_+ = \max(x, 0)$ denotes the hinge function. The overall multi-level loss is formulated as a weighted combination of the three components, governed by the hyperparameters $a$, $b$, and $c$: 
\begin{equation}
\label{equ:Lcon}
    \mathcal{L}_{\text{con}} = a \mathcal{L}_{\text{con}}^{\text{ori}}+b \mathcal{L}_{\text{con}}^{\text{key}}+c \mathcal{L}_{\text{con}}^{\text{unc}}, 
\end{equation}
where $a+b+c=1$.
\subsubsection{Semantic Consistency Constraint:}
To enforce the global semantic consistency of the uncertainty-aware region embeddings, we first compute the average of all region-level representations $\mathbf{U}$ to obtain a global semantic vector for the image. This vector serves as a holistic representation of the image’s semantic content. We then align this global vector with the global mean of the significance-aware patch features $\hat{\mathbf{V}}$, ensuring that the high-level semantic structures across both regions and patches are consistent. The reconstruction loss is defined as follows:
\begin{equation}
\mathcal{L}_{\text{recon}} =   \left\|\frac{1}{K} \sum_{k=1}^{K} \mathbf{U}_k - \frac{1}{L_v} \sum_{l=1}^{L_v} \hat{\mathbf{V}}_l \right\|_2^2.
\end{equation}

\subsubsection{Region Distribution and Diversity Regularization:}
Since the standard normal distribution often serves as an effective model for smooth transformations and complex relationships in latent spaces~\cite{kingma2013vae}, especially in high-dimensional settings, we assume that the prior distribution of the latent variables for the region features follows a standard normal distribution $\mathcal{N}(0, 1)$. To encourage each region's distribution to approximate this prior, we employ the Kullback-Leibler (KL) divergence as a regularization term. Formally,
\begin{equation}
\label{klloss}
\begin{split}
    \mathcal{L}_\text{KL}&=\sum_{k=1}^{K}\text{KL}(\mathcal{N}(\mu_k ,\sigma^2_k )||\mathcal{N}(0,1))\\
    &=-\frac{1}{2}\sum_{k=1}^{K}(1+\log \sigma^2-\mu^2-\sigma ^2 ).
\end{split}
\end{equation}
To prevent attention collapse and promote the diversity of regions, we introduce an entropy regularization term:
\begin{equation}
    \mathcal{L}_\text{ent}=-\frac{1}{K}\sum_{k=1}^{K}\mathbf{A}^k_r\log(\mathbf{A}^k_r),
\end{equation}
where $\mathbf{A}^k_r$ is from equation~\ref{patch-prompt attention}. The overall regularization objective is:
\begin{equation}
\mathcal{L}_{\text{reg}} = \mathcal{L}_{\text{KL}} + \mathcal{L}_{\text{ent}}.
\end{equation}

\begin{table}[t!]
\renewcommand\arraystretch{1}
\setlength{\tabcolsep}{4pt}
\small

\begin{tabular}{l|ccc|ccc|c}
\toprule
\multirow{2}{*}{Settings} & \multicolumn{3}{c|}{Image to Text} & \multicolumn{3}{c|}{Text to Image} & \multirow{2}{*}{rSum} \\
                          & R@1      & R@5      & R@10    & R@1      & R@5      & R@10    &                       \\
                          \midrule
w/o SA                & 73.6   & 93.2   & 96.5  & 60.7   & 86.6   & 92.2  & 502.8                 \\
w/o GA                & 74.9   & 93.8   & 97.0  & 61.5   & 87.3   & 92.5  & 507.0                 \\
w/o RP            & 73.5   & 93.1   & 96.8  & 60.7   & 86.9   & 92.3  & 503.3                 \\
w/o UM          & 74.4   & 93.4   & 97.3  & 62.1   & 88.0   & 93.1  & 508.3  \\
\midrule
Complete & 77.2&	94.5	&97.9&	64.6&	88.5&	93.5	&516.2\\
\bottomrule
\end{tabular}
\caption{The ablation study on the contribution of each module in our proposed method.}
\label{tab:ablation_module}
\end{table}
\section{Experiment}
The datasets, evaluation metrics, and detailed implementation settings used in the experiments are provided in the Appendix.

\subsection{Comparison with State of the Arts}
Following the standard evaluation protocols of the two benchmarks~\cite{faghri2017vse++,zhang2022negative}, we summarize the details of the feature encoders and cross-modal alignment strategies used in all compared methods. Specifically, we include six representative cross-modal alignment approaches~\cite{faghri2017vse++,sgr,scan,chan,laps,avse}, covering a range of alignment paradigms for comprehensive comparison.





\begin{figure}[t]
\centering
\includegraphics[width=1\linewidth]{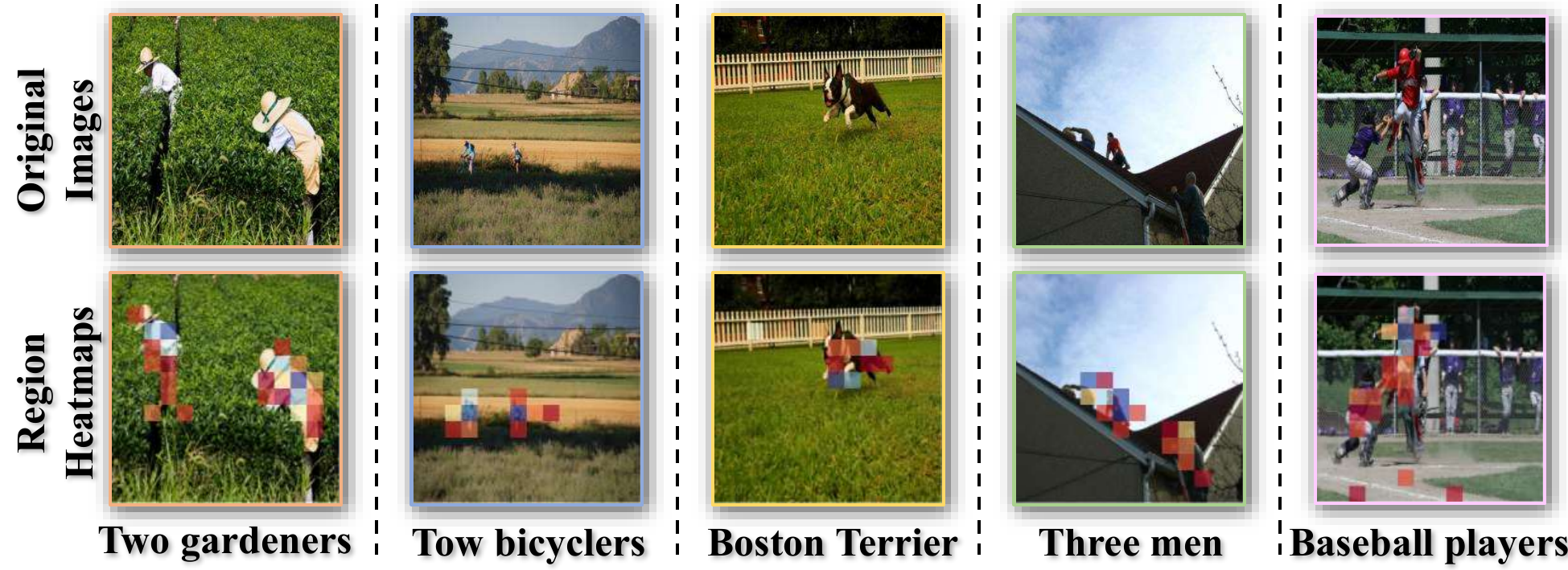}
\caption{The visualization of fine-grained patch-word alignment with each linguistic word.}

\label{fig:patch-word heatmap}
\end{figure}
\begin{figure}[t]
\centering
\includegraphics[width=1.05\linewidth]{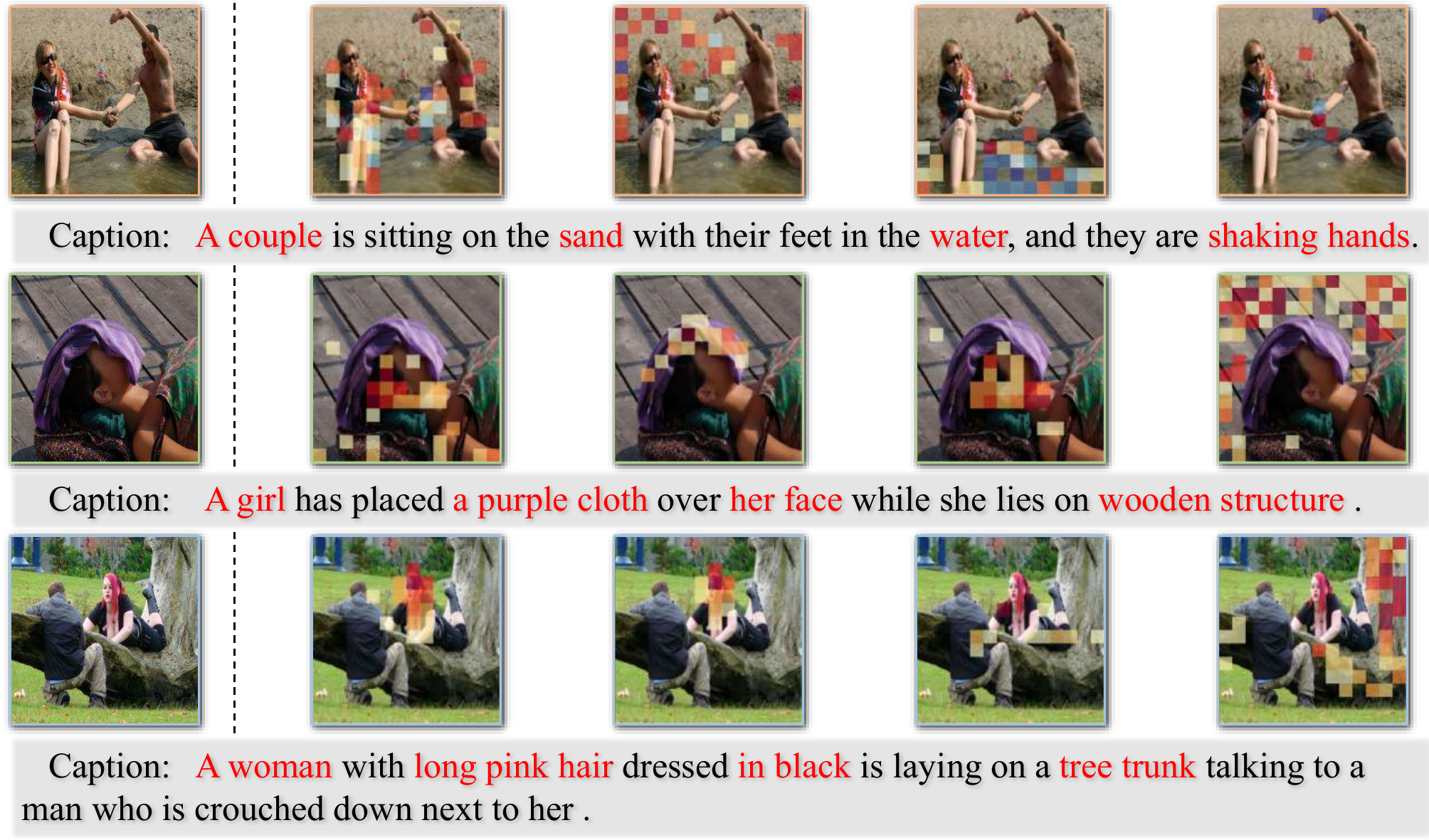}
\caption{The visualization of fine-grained image-text alignment with some key words.}

\label{fig:patch-caption heatmap}
\end{figure}

As shown in Table~\ref{tab:main}, we report the quantitative results on Flickr30K and MS-COCO benchmarks. Our proposed method consistently outperforms all existing state-of-the-art approaches, demonstrating its strong capability in fine-grained cross-modal retrieval. Our method consistently and significantly outperforms all existing state-of-the-art approaches. When evaluated under identical visual and textual encoder configurations, our method demonstrates consistent performance gains in terms of rSum, with improvements ranging from 2.1\%${\sim}$5.6\% on Flickr30K, 1.3\%${\sim}$4.0\% on MS-COCO 1K, and 1.9\%${\sim}$5.6\% on MS-COCO 5K. Notably, while the performance gain narrows slightly as the number of visual patches increases, our approach still maintains a significant margin across all settings.
Compared to two-stage methods that rely on pre-trained object detectors as visual backbones, our approach offers both better performance and greater scalability. More importantly, it eliminates the need for additional detector training and enables end-to-end optimization for downstream tasks, making it more practical and efficient for real-world deployment. To further demonstrate the effectiveness of our proposed method, we provide heatmap visualizations for both patch-word alignment and image-text alignment, as shown in Figure~\ref{fig:patch-word heatmap} and~\ref{fig:patch-caption heatmap}.

\begin{table}[t!]
\renewcommand\arraystretch{1}
\setlength{\tabcolsep}{3.7pt}
\small

\begin{tabular}{l|ccc|ccc|r}
\toprule
\multirow{2}{*}{Settings} & \multicolumn{3}{c|}{Image to Text} & \multicolumn{3}{c|}{Text to Image} & \multirow{2}{*}{rSum} \\
                          & R@1    & R@5    & R@10  & R@1    & R@5    & R@10  &                       \\
                          \midrule
w/o $\mathcal{L}_{\text{con}}^{\text{ori}}$                & 70.1   & 91.3   & 95.7  & 57.6   & 82.2   & 90.3  & 487.2                 \\
w/o $\mathcal{L}_{\text{con}}^{\text{key}}$                & 74.7   & 93.9   & 96.9  & 60.7   & 86.8   & 92.9  & 505.9                 \\
w/o $\mathcal{L}_{\text{con}}^{\text{unc}}$                & 73.8   & 93.1   & 96.1  & 59.9   & 86.1   & 92.6  & 501.6                 \\
w/o $\mathcal{L}_{\text{recon}}$              & 75.8   & 94.2   & 97.8  & 63.1   & 88.5   & 93.8  & 513.2                 \\
w/o $\mathcal{L}_{\text{reg}}$                & 76.2   & 94.7   & 97.5  & 63.5   & 88.5   & 93.8  & 514.2  \\
\midrule
Complete & 77.2&	94.5	&97.9&	64.6&	88.5&	93.5	&516.2\\
\bottomrule
\end{tabular}
\caption{The ablation study on the contribution of each optimization objective in our proposed method.}
\label{tab:ablation_loss}
\end{table}
\subsection{Ablation Studies}

We conduct extensive ablation studies and robustness analyses to evaluate the effectiveness of our approach. All experiments are performed based on the Vision Transformer (ViT-Base) with 224$\times$224 input resolution and BERT-Base as the default text encoder.
\subsubsection{Module Gain.}
To thoroughly assess the contribution of each component, we conduct a module-wise ablation study, as summarized in Table~\ref{tab:ablation_module}. The results demonstrate that both the Significance-aware Adapter (SA) and Region Prompts (RP) are essential components of our method. This performance gap can be attributed to the distribution mismatch between the pretraining corpus and downstream data. The observed improvements underscore the role of SA and RP in mitigating this distribution shift and enhancing model adaptability. Specifically, removing any module leads to a notable performance drop across multiple evaluation metrics, underscoring their individual importance, suggesting that the two components are not only effective in isolation but also highly complementary. This indicates that the design is not merely a sum of independent parts but rather a carefully coordinated structure where each module reinforces the other.

\subsubsection{Optimization Gain.}
To assess the contribution of each loss component, we perform a series of ablation experiments, with the results presented in Table~\ref{tab:ablation_loss}. It is evident that the three losses associated with our Multi-level and Bidirectional Image–Text Alignment—$\mathcal{L}_{\text{con}}^{\text{ori}}$, $\mathcal{L}_{\text{con}}^{\text{key}}$ and $\mathcal{L}_{\text{con}}^{\text{unc}}$
 —play a pivotal role in our method. These objectives directly govern both the alignment of image–text pairs and the modeling of fine-grained uncertainty, thereby exerting a dominant influence on overall performance. In contrast, the loss $\mathcal{L}_{\text{recon}}$ of Semantic Consistency Constraint and the loss $\mathcal{L}_{\text{reg}}$ of Region Distribution and Diversity Regularization serve as auxiliary regularizers: they reinforce the knowledge learned by the region prompts but do not drive the core alignment process. This hierarchy of contributions is corroborated by the ablation results presented in Table~\ref{tab:ablation_loss}, which demonstrate that removing any one of losses leads to the drop in performance, especially the three losses of Multi-level and Bidirectional Image–Text Alignment.
\begin{figure}[t]
\centering
\includegraphics[width=1.0\linewidth]{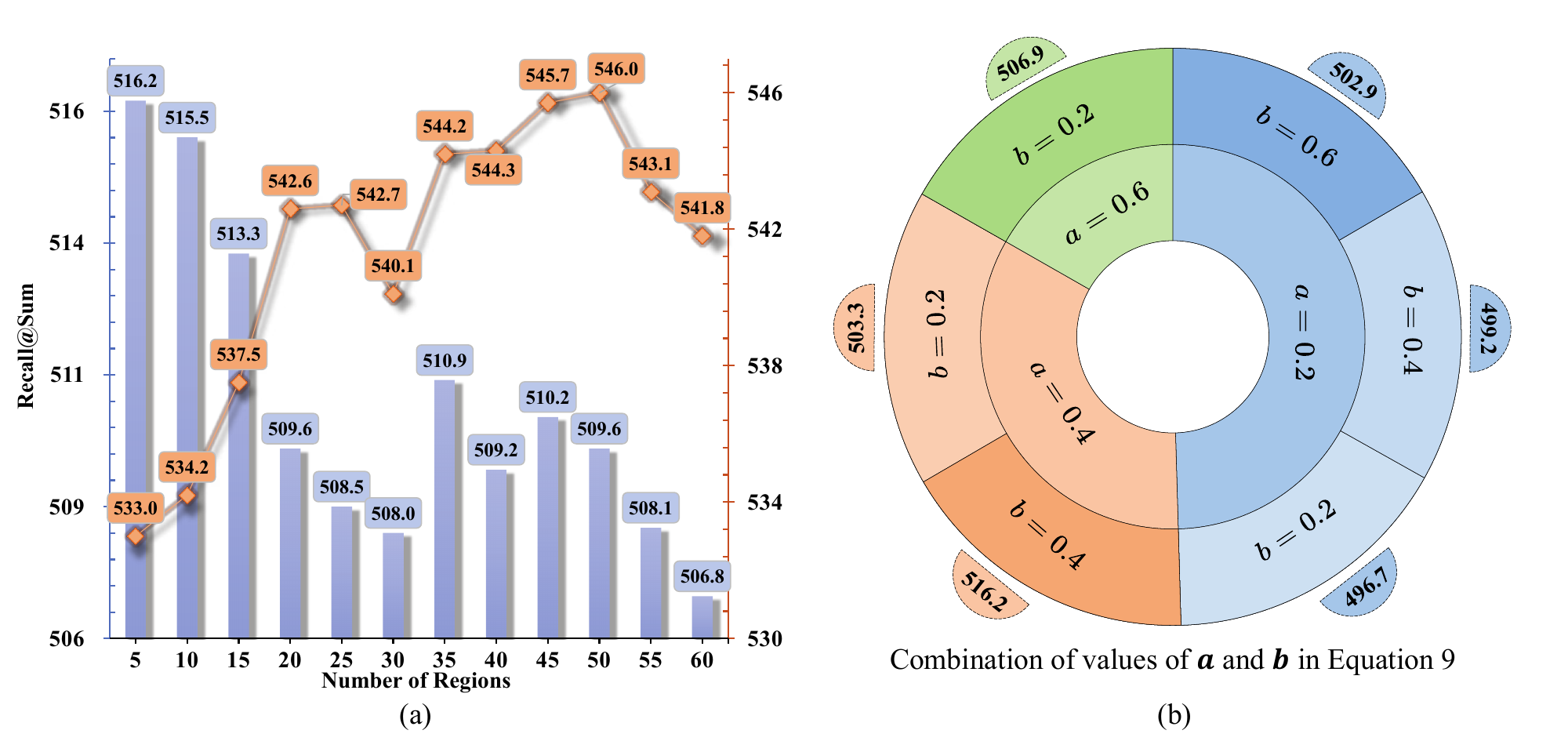}
\caption{
\textbf{Comparison of hyperparameter experimental performance.} The experiments are conducted on the Flickr30k dataset.
\textbf{(a)} Parameter study on the number of region prompts $\mathbf{P}$. The blue bar chart shows results with the visual backbone ViT-base-224, while the yellow line chart shows results with Swin-base-224.
\textbf{(b)} Parameter study on different combinations of weights $a$, $b$ and $c$ in Equation~\ref{equ:Lcon}. Since $a+b+c=1$, only $a$ and $b$ need to be specified. Six combinations are tested: when $a=0.2$ (blue blocks), $b=\{0.2,0.4,0.6\}$; when $a=0.4$ (yellow blocks), $b=\{0.2,0.4\}$; and when $a=0.6$ (green blocks), $b=0.2$.
}

\label{fig:param}
\end{figure}
\subsection{Parameter Analysis}
To investigate the impact of the number of region prompts $\mathbf{P}$ and the weighting proportions in multi-level alignment on model performance, we conducted corresponding hyperparameter experiments. The results are illustrated in Figure~\ref{fig:param}.\\
\textbf{The number of region prompts.} When using ViT-Base-224 as the visual backbone, the optimal performance is achieved when the number of region prompts is set to 5. As the number of prompts increases beyond this point, the retrieval performance exhibits a noticeable decline. In contrast, when adopting Swin-Base-224 as the visual backbone, the best performance is obtained when the number of region prompts reaches 50. We speculate that this discrepancy stems from the architectural differences between the two backbones. Specifically, the Swin Transformer leverages local self-attention and a hierarchical structure, which tends to aggregate local visual information at an earlier stage. As a result, it requires a larger number of learnable region prompts to sufficiently capture and represent the finer-grained semantic prompts embedded in the image. In comparison, ViT’s global attention enables broader contextual modeling from the outset, thereby benefiting more from a smaller set of region prompts.\\
\textbf{Different value combinations of $a$, $b$, $c$.} The results in Figure~
\ref{fig:param}(b) clearly demonstrate that the model performance is highly sensitive to the values of the weighting parameters $a$, $b$, and $c$ in the multi-level alignment. Specifically, the optimal performance is observed when $a$ = $b$ = 0.4 and $c$ = 0.2, while the poorest performance occurs when $a$ = $b$ = 0.2 and $c$ = 0.6. This trend suggests that the original similarity and the significance-aware similarity contribute equally and significantly to the effectiveness of image-text alignment. In contrast, excessive emphasis on uncertainty-aware similarity may introduce noise or overfitting, thereby degrading overall performance. These findings are in line with our ablation results, reinforcing the importance of balancing the contributions from different semantic alignment levels.


\section{Conclusion}
In this paper, we address the critical challenges of fine-grained image-text alignment by identifying two key limitations in existing methods: the lack of effective intra-modal significance modeling and the absence of fine-grained uncertainty quantification. To this end, we propose GRM, a novel and effective approach that introduces new perspectives into cross-modal matching. By leveraging intra-modal semantic biases and modeling region features as a mixture of Gaussian distributions, GRM captures both salient content and region-level uncertainty in a principled manner. This approach enhances alignment robustness, semantic fidelity, and interpretability without over-reliance on global cross-modal attention. Extensive experiments across multiple benchmark datasets and diverse backbone architectures demonstrate the superior and consistent performance of our GRM, offering a strong foundation for future advancements in fine-grained cross-modal understanding and alignment.

\bibliography{aaai2026}
\newpage
\section{Appendix}
\subsection{Datasets and Metrics}
\label{appendix_datasets}
Following prior works~\cite{sgr,faghri2017vse++,scan,laps}, we adopt two widely-used benchmarks—Flickr30K~\cite{flickr30k} and MS-COCO~\cite{mscoco}—for model training and evaluation, where each image is annotated with five corresponding textual descriptions. 
The Flickr30K dataset consists of 29,000 training images, 1,000 test images, and 1,014 validation images. The MS-COCO dataset contains 82,738 training images, 5,000 test images, and 5,000 validation images.
For MS-COCO, we report results in two common settings: (1) the average performance over five-fold cross-validation on 1K test images, and (2) full evaluation on all 5K test images.
Evaluation metrics include Recall@K (\textbf{R@K}), which measures the percentage of correct matches within the top-K retrieved results (K = 1, 5, 10), and \textbf{rSum}, the sum of R@1, R@5, and R@10 for both image-to-text and text-to-image retrieval tasks.

\subsection{Implementation Details}
\label{appendix_implementation}
We adopt Vision Transformer (\textbf{ViT})~\cite{dosovitskiy2020image} with a patch size of 16$\times$16 pixels and Swin Transformer (\textbf{Swin})~\cite{liu2021swin} with a patch size of 32$\times$32 pixels as our visual encoders, and use BERT~\cite{devlin2019bert} as the text encoder. All encoders are based on their standard (base) configurations. The input image resolutions are set to either 224$\times$224 or 384$\times$384, resulting in 14$\times$14 and 24$\times$24 patch tokens for ViT, and 7$\times$7 and 12$\times$12 windows for Swin, respectively.
To ensure compatibility across modalities, we introduce an additional linear projection layer atop each encoder to unify the feature dimension to d = 512.
The entire framework is optimized using the AdamW~\cite{loshchilov2017decoupled} optimizer for 30 epochs, with the margin of the triplet ranking loss set to $\alpha=0. 2$. The number of region prompts $K=5$ for ViT backbones and $K=50$ for Swin backbones. The hyperparameters $a=0.4$, $b=0.4$, $c=0.2$ for both ViT and Siwn backbones. All experiments were performed on one RTX A6000 with 128 batch sizes per epoch.
\end{document}